\documentclass[3p,times,procedia]{elsarticle}
\usepackage{arabtex}
\usepackage{utf8}
\setcode{utf8}
\usepackage{ecrc}
\usepackage[bookmarks=false]{hyperref}
    \hypersetup{colorlinks,
      linkcolor=blue,
      citecolor=blue,
      urlcolor=blue}

\volume{00}

\firstpage{1}

\journalname{Procedia Computer Science}

\runauth{Heakl et al.}

\jid{procs}

\usepackage{amssymb}

\usepackage[figuresright]{rotating}
\usepackage{caption}
\usepackage{subcaption}

\begin{document}
\begin{frontmatter}



\dochead{6th International Conference on AI in Computational Linguistics}%

\title{ArzEn-LLM: Code-Switched Egyptian Arabic-English Translation and Speech Recognition Using LLMs}


\author[a,cor1]{Ahmed Heakl} 
\author[a]{Youssef Zaghloul}
\author[a]{Mennatullah Ali}
\author[b]{Rania Hossam}
\author[a,c]{Walid Gomaa}

\address[a]{Egypt-Japan University of Science and Technology, New Borg El-Arab City, 21934, Alexandria, Egypt}
\address[b]{Mansoura University, Mansoura, 35511, Egypt}
\address[c]{Alexandria University, El-Shatby, 21526, Alexandria, Egypt}

\begin{abstract}

Motivated by the widespread increase in the phenomenon of code-switching between Egyptian Arabic and English in recent times, this paper explores the intricacies of machine translation (MT) and automatic speech recognition (ASR) systems, focusing on translating code-switched Egyptian Arabic-English to either English or Egyptian Arabic. Our goal is to present the methodologies employed in developing these systems, utilizing large language models such as LLama and Gemma. In the field of ASR, we explore the utilization of the Whisper model for code-switched Egyptian Arabic recognition, detailing our experimental procedures including data preprocessing and training techniques. 
Through the implementation of a consecutive speech-to-text translation system that integrates ASR with MT, we aim to overcome challenges posed by limited resources and the unique characteristics of the Egyptian Arabic dialect. Evaluation against established metrics showcases promising results, with our methodologies yielding a significant improvement of $56\%$ in English translation over the state-of-the-art and $9.3\%$ in Arabic translation. Since code-switching is deeply inherent in spoken languages, it is crucial that ASR systems can effectively handle this phenomenon. This capability is crucial for enabling seamless interaction in various domains, including business negotiations, cultural exchanges, and academic discourse. Our models and code are available as open-source resources.
\footnote{Code: \url{http://github.com/ahmedheakl/arazn-llm}}, \footnote{Models: \url{http://huggingface.co/collections/ahmedheakl/arazn-llm-662ceaf12777656607b9524e}}.

\end{abstract}


\cortext[cor1]{Corresponding author. Tel.: +20-1016580028; Email: ahmed.heakl@ejust.edu.eg}
\end{frontmatter}

%

\section{Introduction}
\label{sec: Introduction}

\par 
The term ``code-switching'' describes the phenomenon of a bilingual or multilingual speaker switching between two or more languages~\cite{CSW-Egypt}. It has grown to be a prominent phenomenon in 
multilingual societies around the globe, particularly in the Arab world~\cite{CSW-survey}. In Egypt, code-switching is a significant and common linguistic phenomenon. People's code choices have been impacted by recent political and social changes in Egypt. As shown in Table~\ref{table:translation-examples}, code-switching is evident in everyday conversations.

\par 
Addressing the complexities of code-switching presents a significant challenge due to the vast range of potential data combinations. Compounding this challenge is the scarcity of resources dedicated to training models on code-switched data. Additionally, the extent to which existing language models have encountered code-switched content during pre-training remains uncertain. Consequently, the ability of these models to effectively transfer knowledge to downstream code-switched tasks remains largely unexplored~\cite{CSW-survey}.

\par 
Machine translation approaches include direct-based, which uses dictionaries but lacks analysis~\cite{DBMT}; rule-based, which leverages linguistic rules but requires manual effort~\cite{RBMT}; corpus-based, which relies on data but struggles with low-resource languages~\cite{RBMT}; knowledge-based, which incorporates explicit knowledge but struggles with ambiguity~\cite{KBMT}; and hybrid, which 
combines approaches for better quality~\cite{hybrid-approach}. Arabic and English have different cultural backgrounds, affecting translation. `The news warms my heart' becomes \<الخبر يثلج صدري> in Arabic, where `warms' is translated to \<ثلج> (ices), due to the languages' origins in different climates. This is because English was born in a cold climate, where warmth is a pleasant weather, whereas Arabic was born in a hot climate, where cold is a pleasant weather. Human translators can understand these cultural differences, but machine translators may struggle to capture them \cite{CBMT}. ArzEn corpus serves as a valuable resource for linguistic research and the development of NLP systems capable of handling code-switched Egyptian Arabic-English while preserving cultural aspects~\cite{arzen}.

\begin{table}[h]
    \centering
    \begin{tabular}{|c|c|c|}
        \hline
        Code-switched & English & Egyptian Arabic \\
        \hline
        \RL{\LR{meeting} فى الشركة} & meeting at the company & \RL{اجتماع فى الشركة} \\
       \RL{نجرب اكل \LR{italian}} & try Italian food & \RL{نجرب اكل ايطالى} \\
      \RL{اعمل \LR{check} لل\LR{email}} & I check the email & \< ببص على رسايلي> \\
    
        \hline
    \end{tabular}
    \caption{Examples of English and Egyptian Arabic human translations.}
    \label{table:translation-examples}
\end{table}

\par 
Our primary contributions are the following:

\begin{itemize}
    \item Translation: Developing translation models using open-source models (Llma2, Llama3, and Gemma) for code-switched Egyptian Arabic-English, aiming to achieve translations that closely mimic human-generated outputs, from code-switched Egyptian Arabic to either English or Egyptian Arabic.
    \item ASR: Developing an Automatic Speech Recognition (ASR) system using Whisper as a crucial component of a complete pipeline, where spoken code-switched Egyptian Arabic-English utterances are transcribed into written text, which is then translated using machine translation.
    \item Quantization: Quantizing our models to be more accessible to human users through their CPUs/GPUs, ensuring efficient deployment and utilization of our models
    \item Evaluation framework: Extending available metrics to enhance the reliability of our models, prioritizing evaluation accuracy and performance.
    \item Open-Sourcing: Making our models and code publicly available to encourage community engagement and further research.
\end{itemize}

\par 
The rest of this paper is organized as follows. Section \ref{sec: Related Works} reviews related literature. Section \ref{sec: Methodology} gives our methodology and experimental work. Section \ref{sec: Results} presents our results and discussion, featuring evaluations across multiple metrics. Concluding remarks are provided in Section \ref{sec: conclusion}.

\section{Related Works}
\label{sec: Related Works}

\subsection{Enhancements in Code-Switching Resources for Egyptian Arabic}


\par 
The authors in~\cite{CSW-Egypt} discussed the phenomenon of code-switching (CSW) in Egyptian movies where code-switching is prevalent due to the complex linguistic landscape and social variables~\cite{CSW-worldwide}, where speakers seamlessly blend dialectal Egyptian Arabic with other languages like English and French. 
The authors in~\cite{arzen-threeway} introduced ArzEn-ST which is a three-way speech translation corpus for code-switched Egyptian Arabic-English, which extends the ArzEn corpus~\cite{arzen}.  
They also presented benchmark baseline results for ASR, MT, and speech translation (ST) tasks. 
In addition, the authors in ~\cite{arzen-parallel} expanded the existing Egyptian Arabic datasets by introducing a new dataset focused on daily life conversations from movies and songs. This dataset is designed for benchmarking new machine translation models, fine-tuning large language models in few-shot settings, and facilitating research in cross-linguistic analysis and lexical semantics. This also helps in capturing more cultural nuances related to Egyptian Arabic.



\subsection{Code-switched corpora}

\par 
The authors in~\cite{arzen} presented the ArzEn corpus, an Egyptian Arabic-English code-switching spontaneous speech corpus. The corpus comprises 12 hours of recorded interviews with 38 Egyptian bilingual university students and employees. The corpus is designed for Automatic Speech Recognition (ASR) systems and offers insights into linguistic, sociological, and psychological aspects of code-switching. The work done in~\cite{arzen-threeway} extends the ArzEn corpus with translation in both primary (Egyptian-Arabic) and secondary (English) languages. 
The authors in~\cite{arzen-parallel} presented ArzEn-MultiGenre corpus comprising 25,557 segment pairs of Egyptian Arabic song lyrics, novels, and TV show subtitles, all manually translated and aligned with their English counterparts.

\subsection{The era of Large Language Models (LLMs)}

\par 
The process of translation requires a complete understanding of linguistic conversion, syntactic, grammatical, and cultural dimensions. It is more than mapping words between languages~\cite{llm-intro}.  Accurate translation requires a deep understanding of the cultural nuances inherent in both languages, ensuring the preservation of cultural sensitivity and local values~\cite{acegpt}. 
This versatility of LLMs enabled them to excel in numerous NLP applications, such as text generation (Llama2~\cite{llama2}, ChatGPT~\cite{chatgpt}), machine translation (NLLB~\cite{NLLB}, SemalessM4T~\cite{seamlessm4t}, ArzEn-ST~\cite{arzen-threeway}). 
Recent advancements in Large Language Models (LLMs) have led to the development of powerful models like LLaMa2~\cite{llama2}, Gemma (2B, 7B)~\cite{gemma}, and LLaMa3 8B, which have demonstrated impressive capabilities in NLP tasks. 
Notably, these models have been designed to be more computationally efficient, allowing them to be deployed on consumer-grade GPUs. This shift enables researchers and developers to harness the power of LLMs on local machines, facilitating faster experimentation, prototyping, and deployment of AI applications.

\subsection{Code-switching Automatic Speech Recognition (CSW-ASR)}

\par 
Researchers have explored acoustic, linguistic, and pronunciation modeling approaches, including language identification systems~\cite{ASR-1}, parallel recognizers~\cite{ASR-2}, and single-pass 
methods~\cite{ASR-3}. The authors in~\cite{whisper} presented Whisper, a speech recognition system trained on 680,000 hours of multilingual and multitask audio data, achieving zero-shot transfer 
capabilities and approaching human accuracy and robustness. The system's architecture is based on an encoder-decoder transformer, leveraging a minimalist data processing approach and multitask training.

\section{Methodology}
\label{sec: Methodology}

\par 
In this section, we present the machine translation and automatic speech recognition systems we used. 

\subsection{Machine Translation (MT)}

\par 
The task of machine translation is represented by a mapping $\mathcal{T}: X^{S} \rightarrow Y^{T}$ where $\mathcal{T}$ is the machine translation function, $X^{S}$ is the set of source sentences in the
source language $S$, represented as a sequence of tokens $x = (x_1, x_2, ..., x_n)$,  
and $Y^{T}$ is the set of translated sentences in the target language $T$. 
The goal of machine translation is to find the optimal translation $\hat{y}$ that maximizes the likelihood of the target sentence given the source sentence $\hat{y} = \arg\max_{y \in Y^{T}} P(y|x)$ where $P(y|x)$ is the conditional probability of the target sentence given the source sentence. Formally, we can define the machine translation problem as:
\begin{equation}\label{eq:mt-opt}
\centering
\mathcal{T}^* = \arg\min_{\mathcal{T}} \mathbb{E}_{x \sim \mathcal{X}^{\mathcal{S}}} [d(\mathcal{T}(x), y^*)]
\end{equation}

where $\mathcal{T}^*$ is the optimal machine translation function. $d(\cdot, \cdot)$ is a distance metric (e.g. BLEU score, METEOR score) that measures the similarity between the translated sentence and the reference translation $y^*$. The goal is to find the optimal machine translation function $\mathcal{T}^*$ that minimizes the expected distance between the translated sentence and the reference translation. This mathematical definition provides a formal framework for understanding the task of machine translation and its optimization problem.

\par 
We used the infamous translation ArzEn-ST dataset to train all of our models~\cite{arzen-threeway}. We adhere to the same train and test splits as described in~\cite{arzen}. 
Specifically, we utilize the ArzEn-ST test set, comprising 1,402 sentences, and the train set, consisting of 3,344 sentences. To provide our models with a richer context, we also pre-train them on larger datasets, including the entire parallel corpora presented in~\cite{arzen-parallel}. This approach enables our models to leverage a broader range of linguistic patterns and cultural nuances.

\par
Data pre-processing involves removing corpus-specific annotations, URLs, and emoticons, as well as converting 
all text to lowercase. This step is crucial in ensuring that our models focus on the underlying linguistic structures and cultural nuances of the Egyptian-Arabic language.

\par
Given the sequential nature of the translation task and the need for culturally enriched translations, we opt for large language models (LLMs) as our primary approach. Specifically, we employ the latest LLMs that can be accommodated by consumer-grade RAM or GPU, including LLaMA3 8B, Gemma1.1 2B, and Gemma1.1 7B~\cite{gemma}. Notably, we utilize the chat version of each model, which has been trained to follow human instructions, thereby facilitating the training process. 
All models are trained using 2 T4 GPUs with 16GB VRAM. It is worth noting that these models are decode-based architectures, which are particularly well-suited for sequential tasks like machine translation. By leveraging the strengths of these models, we aim to produce culturally fitting translations that capture the nuances of Egyptian-Arabic language and culture.

\par 
We employed the paged-Adam optimizer with weight decay~\cite{adamw} in 32-bit precision for all models, except for LLaMa3, which required 8-bit precision due to its substantial size (8 billion parameters). To accommodate the computational demands of the Adam optimizer, which utilizes multiple gradient copies, we trained our models using adapters for LLMs. Specifically, we explored the use of Quantized low-Rank Adapters (QLoRA)~\cite{qlora} and weight-Decomposed low-Rank Adaptation (DoRA)~\cite{dora}, with the latter yielding the most promising results and exhibiting similar behavior to the original fine-tuning process. We opted for int4 quantization with normal floats (nf4) for each adapter.

\par 
To mitigate memory constraints during training, we leveraged gradient checkpointing~\cite{gradient-checkpoint}, which incurs only an additional forward pass per mini-batch, while reducing memory consumption to $O(\sqrt{n})$. Furthermore, to enable training with effectively large batch sizes while minimizing memory constraints, we implemented a gradient accumulation step of 4~\cite{gradient-accumulation}. This approach allows us to accumulate gradients from 4 batches, perform backward propagation, and achieve comparable accuracy to updating a batch of 4 at once, while reducing memory requirements by a factor of 4.

\par 
Our experiments revealed that the optimal strategy involves training models for a single epoch with a constant learning rate schedule. Additionally, we ensured that input attention masks were configured to mask out the output translation, thereby computing gradients and loss only for the output translation. Lastly, to make our models available on a consumer CPU, we provide the quanitized GGUF version of our best model. The quantization was done through the implementation of GGUF llama.cpp. 

\subsection{Automatic Speech Recognition (ASR)}

\par 
In the context of Automatic Speech Recognition (ASR), we aim to convert a speech signal into a sequence of words. Let's assume a speech signal $x = (x_1, x_2, ..., x_T)$ where $x_t \in \mathbb{R}^D$ is the acoustic feature vector at time $t$ and $T$ is the length of the speech signal. The goal of ASR is to find the most likely sequence of words $w = (w_1, w_2, ..., 
w_N)$ where $w_n \in \mathcal{V}$ is the $n^{th}$ word in the vocabulary $\mathcal{V}$ and $N$ is the length of the transcription. The ASR problem can be formulated as $\hat{w} = \arg\max_{w \in \mathcal{V}^*} P(w|x)$, where $P(w|x)$ is the 
posterior probability of the word sequence $w$ given the speech signal $x$.

\par 
We propose a cascaded speech-to-text translation system, wherein an ASR system is trained to generate transcriptions, which are subsequently fed into a machine translation  model. We opted for a cascaded architecture over an end-to-end approach due to the constraints imposed by limited resources, which rendered the development of an end-to-end system infeasible. Furthermore, previous research has demonstrated that cascaded systems can outperform end-to-end systems in low-resource settings, thereby motivating our design choice~\cite{cascaded-mt}

\par 
We employed the Whisper model~\cite{whisper} to tackle the task of ASR for Egyptian Arabic. The Whisper model, trained on a large-scale dataset of 680,000 hours of multilingual and multitask supervision, demonstrated excellent generalizability to our specific use case. This is particularly valuable for our application, as we are dealing with a unique dialect of Arabic, namely the Egyptian Arabic. The Whisper model, an encoder-decoder architecture, takes the input signal in spectrogram format and utilizes cross-attention mechanisms. For our experiments, we leveraged the ArzEn-ST dataset~\cite{arzen}, but restricted the output to transcription only, focusing on code-switched Egyptian Arabic.

\par 
Data preprocessing involved resampling all audio to 16 kHz, removing URLs and emoticons from the text, segmenting the speech into 30-second clips, and converting each clip into mel-spectrogram images. Training was conducted on 2 T4 GPUs, each equipped with 16 GB of VRAM. The training process was completed in approximately 5 hours.

\begin{table}[t]
\centering
\begin{tabular}{|l|c|c|c|c|c|c|c|}
\hline
\textbf{Model} & \textbf{BLEU $\uparrow$} & \textbf{BERT-F1 $\uparrow$} & \textbf{EED $\downarrow$} & \textbf{METEOR $\uparrow$} & \textbf{LLMG $\uparrow$}\\
\hline
Hamed et al., 2022 \cite{arzen-threeway} & 8.6 & - & - & - & -\\ 
Hamed et al., 2022 + Extra & 34.3 & - & - & - & - \\ \hline
LLaMa2 7B & 26.2 & 42.9\% & 0.68 & 0.12 & 48\% \\
Gemma1.1 2B & 34.3 & 72.1\% & 0.41 & 0.39 & 75.8\% \\
Gemma1.1 2B + Extra & 37.5 & 75.8\% & 0.37 & 0.56 & 79.6\% \\
Gemma1.1 7B & 38 & 77.0\% & 0.37 & 0.53 & 84.6\% \\
Gemma1.1 7B + Extra & 38.2 & 77.6\% & 0.37 & 0.56 & 84.3\% \\
LLaMa3 8B GGUF Q5 & 53.01 & 80.8\% & 0.31 & 0.58 & 86.2\% \\
LLaMa3 8B & \textbf{53.64} & \textbf{81.1\%} & 0.31 & \textbf{0.62} & \textbf{86.4\%} \\
LLaMa3 8B + Extra & 52.27 & 80.1\% & \textbf{0.30} & 0.59 & 85.8\% \\
\hline
\end{tabular}
\caption{Summary results for the models trained on ArzEn-ST to generate English translations. We report BLEU score using SacreBLEU~\cite{bleu}, 
BERT F1, Edit Distance (EED), METEOR, and LLaMa3 70B as an LLM Grader (LLMG). The lower section of the table represents our work.}
\label{table:english-results}
\end{table}

\begin{table}[t]
\centering
\begin{tabular}{|l|c|c|c|c|c|c|c|}
\hline
\textbf{Model} & \textbf{BLEU $\uparrow$} & \textbf{BERT-F1 $\uparrow$} & \textbf{EED $\downarrow$} & \textbf{METEOR $\uparrow$} & \textbf{LLMG $\uparrow$}\\
\hline
Hamed et al., 2022 \cite{arzen-threeway} & 48.0 & - & - & - & - \\ 
Hamed et al., 2022 + Extra & 79.8 & - & - & - & - \\ \hline
Gemma1.1 2B & 86.9 & 97.1\% & 0.09 & 0.87 & 94\% \\
Gemma1.1 7B & 83.7 & 95.9\% & 0.12 & 0.84 & 92.6\% \\
LLaMa3 8B GGUF Q5 & 86.3 & 96.2\% & 0.09 & 0.76 & 94\% \\
LLaMa3 8B & \textbf{87.2} & \textbf{98.8\%} & \textbf{0.07} & \textbf{0.88}  & \textbf{96\%} \\
\hline

\end{tabular}
\caption{Summary results for the models trained on ArzEn-ST to generate Egyptian Arabic translations. We report BLEU score using SacreBLEU~\cite{bleu}, BERT F1, Edit Distance (EED), METEOR, and LLaMa3 70B as an LLM Grader (LLMG). The lower section of the table represents our work.}
\label{table:arabic-results}
\end{table}

\section{Results and Discussion}
\label{sec: Results}

\par 
We evaluated the machine translation models using five criteria: BLEU~\cite{bleu}, BERT Score~\cite{bert-score}, edit distance (EED), METEOR~\cite{meteor}, and LLaMa3-based grading, inspired by~\cite{llm-grader}, as traditional metrics are limited in capturing semantic nuances. For ASR, we employed Word-Error Rate (WER) and Character-Error Rate (CER) as evaluation metrics. Our models are compared to the state-of-the-art results in~\cite{arzen-threeway}, with a focus on BLEU for MT and WER and CER for ASR, as these are the only reported metrics.

\begin{figure}[t]
    \centering
    \begin{subfigure}[b]{0.49\textwidth}
      \includegraphics[width=\linewidth]{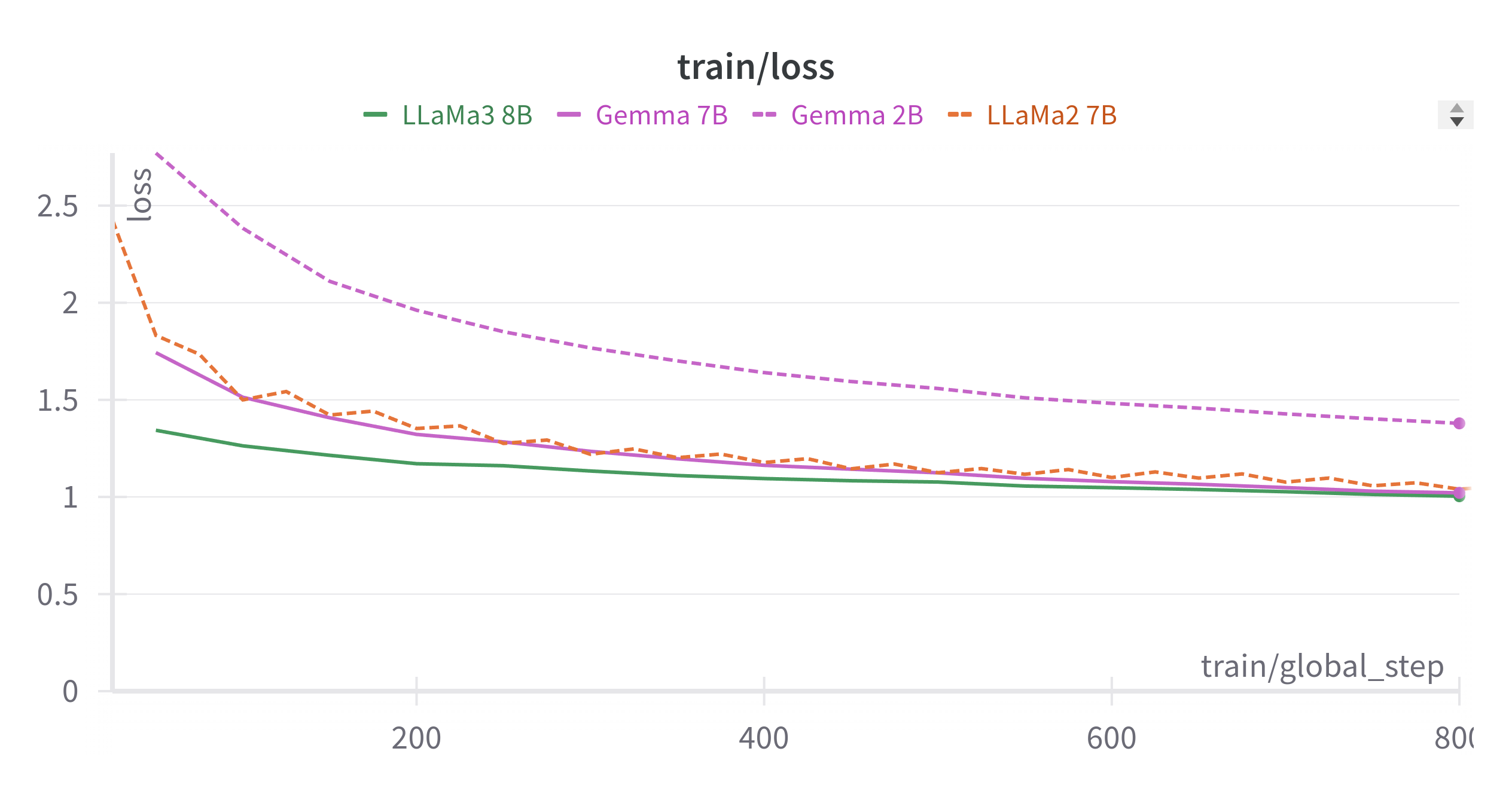}\par
      \caption{Training curves for English translation training.}
      \label{fig:train-eng} 
    \end{subfigure}
    \hfill
    \begin{subfigure}[b]{0.49\textwidth}
      \includegraphics[width=\linewidth]{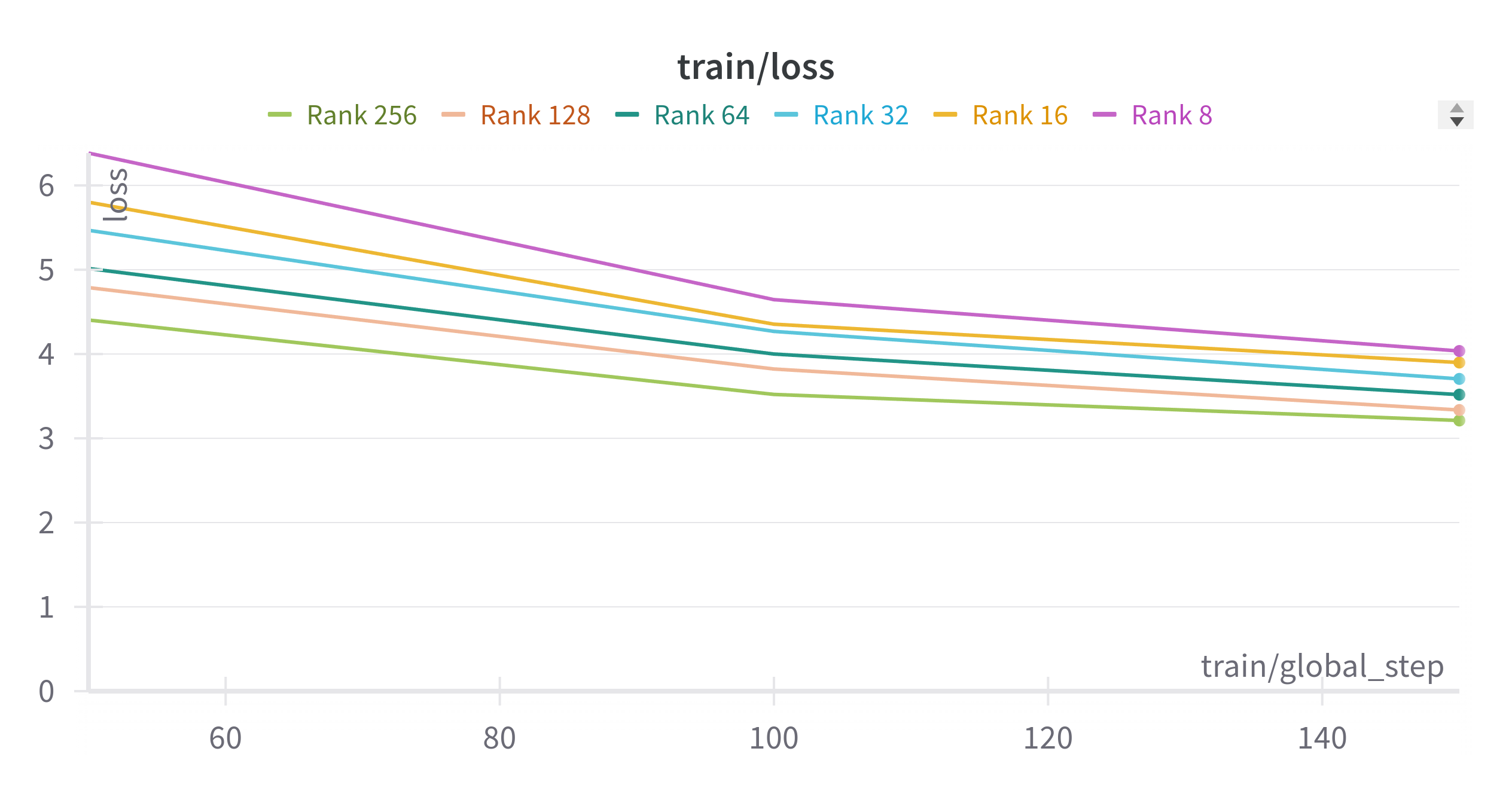}\par
      \caption{Training curves for different QLoRA ranks on Gemma1.1 2B.}
      \label{fig:train-ranks} 
    \end{subfigure}
    \hfill
    \begin{subfigure}[b]{0.49\textwidth}
      \includegraphics[width=\linewidth]{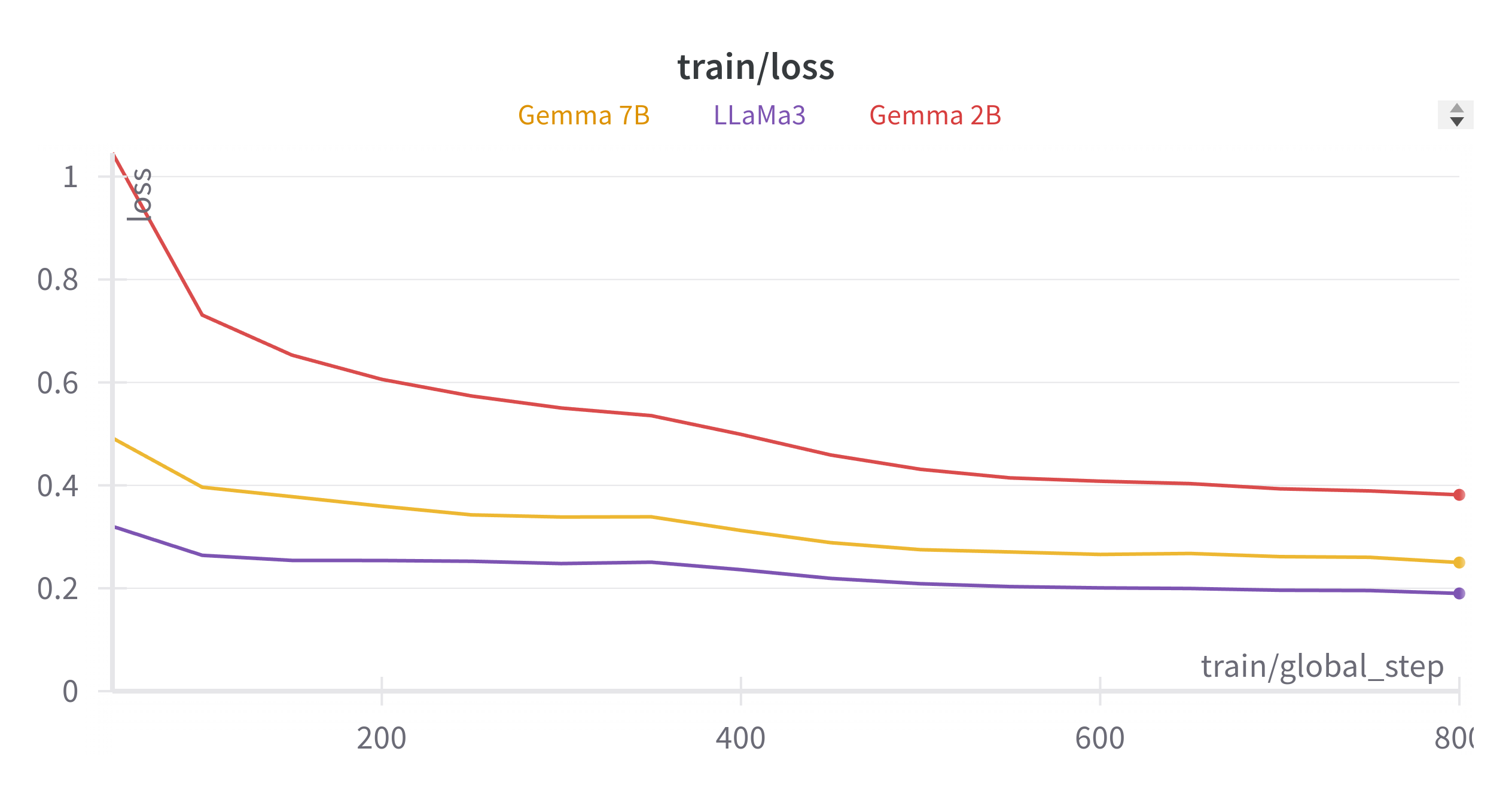}\par
      \caption{Training curves for Arabic translation training.}
      \label{fig:train-ar} 
    \end{subfigure}
    \hfill
    \begin{subfigure}{0.49\textwidth}
      \includegraphics[width=\linewidth]{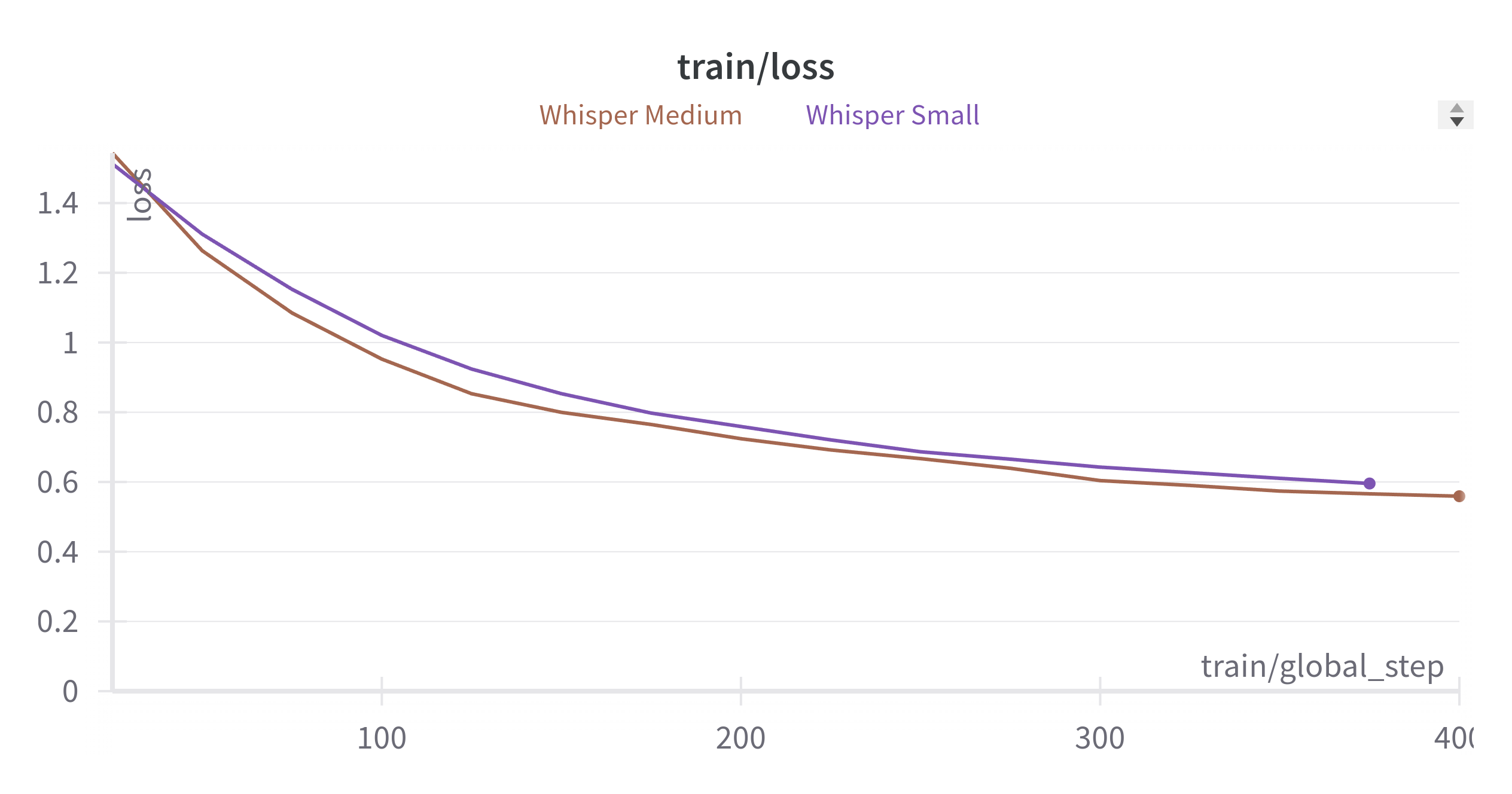}\par
      \caption{Training curves for Whisper.}
      \label{fig:train-whisper} 
    \end{subfigure}
    \caption{Training curves for various machine translation models. The x-axis represents the number of training steps, and the y-axis represents the loss. }
    \label{fig:training-curves}
\end{figure}

Figure~\ref{fig:train-eng} shows that LLaMa3 outperforms all other models on the ArzEn to English translation task. As in table~\ref{table:english-results}, LLaMa3 achieves a BLEU score of $53.64$, which is significantly higher than the SoTA~\cite{arzen-threeway} by $56\%$. Also, smaller models such as Gemma 2B and Gemma 7B achieved comparable results to LLaMa3 8B with $9\%$ and $4.1\%$ lower in BERT-f1 score, respectively.
On the other hand, LLaMa2 performance is the lowest which can be easily interpreted due to the fact that its tokenizer does not support Arabic tokenization. In contrast to new models such as Gemma and LLaMa3 which uses Byte-Pair Encoding (BPE)~\cite{bpe} implemented with tiktoken, LLaMa2 just breaks down the Arabic sentence into characters as shown in table~\ref{table:tokenizers-comparison}. 

Notably, models pre-trained on additional data (Hamed et al., 2022~\cite{arzen-threeway} + Extra and Gemmal.1 2B + Extra) generally outperform their counterparts trained only on the ArzEn dataset, suggesting that extra pre-training data can effectively enhance machine translation model performance. Although, this gain is marginal for larger models, such as Gemma1.1 7B, it can even be detrimental, as observed in LLaMa3 8B, with a $-1\%$ decrease in BERT-f1 score.

\begin{table}[b]
\centering
\begin{tabular}{|l|c|c|c|c|c|}
\hline
\textbf{Model} & \textbf{WER $\downarrow$} & \textbf{CER $\downarrow$} & \textbf{BLEU $\uparrow$} & \textbf{LLMG $\uparrow$} & \textbf{EED $\downarrow$} \\
\hline
Hamed et al., 2022 \cite{arzen-threeway} & 57.9 & 36.2 & - & - & - \\ 
Hamed et al., 2022 + Extra & 34.7 & 20.0 & - & - & -  \\ \hline
Whisper Small & 32.6 & 12.8 & 51.77 & 88.1\% & 0.14 \\
Whisper Medium & \textbf{31.1} & \textbf{12.0} & \textbf{55.41} & \textbf{92.5\%} & \textbf{0.09} \\
\hline
\end{tabular}
\caption{Performance of automatic speech recognition from speech to code-switched Arabic-English task. Lower Word-Error Rate (WER), Character-Error Rate (CER), and Edit Distance (EED) scores indicate better quality. The lower section of the table represents our work.}
\label{table:asr-results}
\end{table}

As shown in table~\ref{table:arabic-results}, translating into Arabic yields significantly higher BLEU scores compared to translating into English, with our optimal Arabic model achieving a BLEU score of 87.2, whereas the best English model attains a BLEU score of 53.64, representing a notable difference of approximately 62\%. This phenomenon is consistent with the linguistic characteristics of the source text, where a significant proportion (approximately $85\%$) of Arabic words remain largely unchanged, with only minor modifications required to accommodate the target language.

As illustrated in figure~\ref{fig:train-ranks}, increasing the LoRA rank consistently yields better results. Our experiments reveal that the optimal parameters are a rank of 256 and an alpha value of 128. Furthermore, we observe that higher ranks require increased LoRA dropout to mitigate overfitting, with a dropout of 0.1 employed for ranks exceeding 32.

As shown in table~\ref{table:asr-results}, our trained Whisper models surpass the state-of-the-art results in~\cite{arzen-threeway} (+ Extra) by $11.6\%$ in WER, despite being trained solely on the original data without additional pre-training. Furthermore, figure~\ref{fig:train-whisper} illustrates that the medium Whisper model marginally outperforms the small version, resulting in a $7.1\%$ increase in BLEU score, as reflected in table~\ref{table:asr-results}. Whisper can achieve real-time output, with a latency of $1.3$ seconds for a single 30-second clip inference on a consumer-grade GPU with fp16 precision, and 18 seconds on a CPU.
\par 
Notably, for English models, we found that human evaluation is particularly well-suited. Therefore, we conducted a human evaluation study, where 65 university students were asked to assess the quality of 10 randomly selected generated sentences on a scale of 1-10, with 1 indicating an irrelevant translation and 10 representing a perfect translation that captures both meaning and cultural nuances. This approach was necessary, as traditional evaluation metrics such as BERTScore, METEOR, edit distance, and BLEU fail to adequately capture the nuances of meaning and cultural context. Our results show that, on average, the generated translations received a rating of 9.2 out of 10, which supports our claim of capturing both perfect meaning and cultural nuances. For instance, when presented with the sentence ``\RL{انا دخلت \LR{IG}},'' our model produced the translation "I entered IG school," notwithstanding that ``IG'' signifies ``Instagram'' 
in contexts outside of Egyptian culture.

\begin{table}[t]
\centering
\begin{tabular}{|l|r|}
\hline
\textbf{Model} & \textbf{Tokens} \\
\hline
LLaMa2 & [\<أ، ن، ا، أ، ح، ب، ا، ل، ت، ف، ا، ح>] \\
Gemma  & [\<أنا، أح، ب، التف، اح>] \\
LLaMa3  & [\<أنا، أح، ب، التف، اح>] \\
\hline
\end{tabular}
\caption{Tokenization results for the Arabic sentence \RL{أنا أحب التفاح} (I love apples) using different tokenizers. }
\label{table:tokenizers-comparison}
\end{table}

Finally, our top-performing model, LLaMa3, was quantized from bfloat16 to 5-bit Q5, achieving a 68.75\% reduction in bits while maintaining performance, with only 1.2\% and 1\% degradation in English and Arabic versions, respectively. The quantized model can be deployed on a consumer-grade RAM with a modest 5.6 GB footprint, supporting a throughput of 7.2 tokens/sec, thereby enabling real-time speech translation and video dubbing applications.
\section{Conclusion}
\label{sec: conclusion}
This paper has provided insights into the methodologies employed in developing machine translation and automatic speech recognition systems for code-switched Egyptian Arabic. Through careful experimentation and rigorous evaluation, we have demonstrated the effectiveness of our approaches in achieving culturally fitting translations and accurate speech recognition.
\par 
Our findings emphasize the importance of using large language models and pre-training with additional data to enhance the performance of MT systems. Moreover, the success of our ASR models, particularly the Whisper architecture, highlights the potential of deep learning techniques in tackling speech recognition tasks, even in low-resource settings.
\par 
Looking ahead, further research could explore advanced optimization techniques and novel model architectures to push the boundaries of MT and ASR performance. Additionally, efforts to expand training data and refine models for specific dialects could result in even more precise translations and transcriptions, fostering greater linguistic accessibility in our globalized world.

\end{document}